\documentclass[12pt,a4paper]{article}

\usepackage[margin=0.8in]{geometry}
\usepackage{setspace} 
\usepackage{authblk}
\usepackage{times}
\usepackage{cite}
\usepackage{graphicx}
\usepackage{subfig}
\usepackage{upgreek}
\usepackage{amsthm,amsmath,amssymb,bm}
\usepackage{CJKutf8} 
\usepackage{enumitem}
\usepackage{xurl}
\usepackage{color}
\usepackage[dvipsnames]{xcolor}
\usepackage[hidelinks,colorlinks=false]{hyperref}
\renewcommand{\figurename}{Fig.}

\graphicspath{{figures/}}

\title{\textbf{GenSwarm: Scalable Multi-Robot Code-Policy Generation and Deployment via Language Models}}

\author[1]{Wenkang Ji}
\author[1]{Huaben Chen}
\author[1]{Mingyang Chen}
\author[2]{Guobin Zhu}
\author[3]{Lufeng Xu}
\author[4,5]{\\Roderich Gro{\ss}}
\author[2]{Rui Zhou}
\author[3]{Ming Cao}
\author[1*]{Shiyu Zhao}

\affil[1]{Department of Artificial Intelligence, Westlake University, Hangzhou, China}
\affil[2]{School of Automation Science and Electrical Engineering, Beihang University, Beijing, China}
\affil[3]{Institute of Engineering and Technology, University of Groningen, Groningen, Netherlands}
\affil[4]{Department of Computer Science, Technical University of Darmstadt, Darmstadt, Germany}
\affil[5]{School of Electrical and Electronic Engineering, The University of Sheffield, Sheffield, UK}
\affil[*]{To whom correspondence should be addressed; E-mail: zhaoshiyu@westlake.edu.cn.}

\date{}

\begin{document}
\begin{CJK}{UTF8}{gbsn} 

\maketitle

\begin{abstract}
The development of control policies for multi-robot systems traditionally follows a complex and labor-intensive process, often lacking the flexibility to adapt to dynamic tasks. This has motivated research on methods to automatically create control policies. However, these methods require iterative processes of manually crafting and refining objective functions, thereby prolonging the development cycle. This work introduces \textit{GenSwarm}, an end-to-end system that leverages large language models to automatically generate and deploy control policies for real-world multi-robot systems based on user instructions in natural language. As a multi-language-agent system, GenSwarm achieves zero-shot learning, enabling rapid adaptation to altered or unseen tasks. The white-box nature of the code policies ensures strong reproducibility and interpretability. With its scalable software and hardware architectures, GenSwarm supports efficient and automated policy deployment on both simulated and real-world multi-robot systems, realizing an instruction-to-execution end-to-end functionality that may transform the development paradigm of multi-robot systems in the future.
\end{abstract}
\newpage

\section*{Introduction}
Multi-robot systems show significant promise for applications both indoors (for example, factory floors, warehouses, hospitals) and outdoors (for example, transport, inspection, farming, disaster response) \cite{dorigo2021swarm,marques-dars22}. The present paradigm of developing multi-robot systems follows a complex and labor-intensive process that involves steps like task analysis, algorithm design, code programming, simulation validation, and real-world deployment. This paradigm requires skilled professionals who are familiar with both theories and software/hardware implementation, incurring high costs in human resources. Moreover, it does not adapt well to dynamically changing tasks: the emergence of a new task requires the repetition of the complex process.

Automatic generation and deployment of control policies for multi-robot systems is an appealing paradigm, as it promises substantial savings in terms of human effort and other resources \cite{francesca2016automatic,lopes2016supervisory,hasselmann2021empirical}. However, this paradigm is nontrivial to realize as a multi-robot group as a whole cannot be programmed directly; rather, a desired collective behavior can be achieved only by programming each individual robot, which relies on its locally available information. Previous methods for automatic development of multi-robot systems are primarily based on optimization techniques \cite{francesca2016automatic,hasselmann2021empirical}. For instance, an objective function is first crafted to mathematically describe a desired task and then optimized to generate policies through methods such as evolutionary computation \cite{francesca2014automode,Bredeche-Frontiers2018,hasselmann2021empirical} or systematic search \cite{gauci2014self}.
Despite their promise, these optimization methods face the common limitation of requiring manual crafting of objective functions.

Recent advances in large language models (LLMs) \cite{brown2020language,chen2021evaluating} and vision language models (VLMs) \cite{zhang2024vision,hurst2024gpt} offer new paradigms for developing robotic systems. In one paradigm, a language model can be deployed onboard a robot to directly make decisions online \cite{xu2023exploring,ma2023large,li2024challenges,chen2024solving}. Due to the generality of language models, this paradigm could be used to address open-ended tasks \cite{ahn2022can,park2023generative,zheng2023steve}. However, it faces challenges in terms of reproducibility, interpretability, and hallucination. In another paradigm, a language model is used to generate executable code policies that are subsequently uploaded for execution on-board robots. A representative method that falls into this paradigm is Code-as-Policy (CaP) \cite{liang2023code, singh2023progprompt,xu2023creative}. Due to the white-box nature of executable code, this paradigm offers high reproducibility and interpretability. Moreover, since executable code usually requires fewer resources than LLMs, this paradigm also enables real-time control on low-cost robot platforms. This is especially relevant for large-scale multi-robot systems, where collective behaviors emerge from robots with exceedingly limited onboard resources \cite{rubenstein2014programmable,vasarhelyi2018optimized,slavkov2018morphogenesis,berlinger2021implicit,zhou2022swarm,sun2023mean}.
Therefore, this code-policy paradigm is adopted in our work.

Despite the promise of the code-policy paradigm, the development of control policies 
for multi-robot systems faces additional challenges
compared to single-robot systems \cite{liang2023code,vemprala2024chatgpt,jin2024robotgpt}. First, the design of policies must consider a robot's interactions with its peers. In some situations, the robot may compete with its peers, for example, for limited resources, whereas in others it may cooperate with its peers to achieve a common goal \cite{sun2023mean,wang2020shape,hasselmann2021empirical}.
Second, the deployment and maintenance of policies require scalable software and hardware systems, which is particularly relevant for multi-robot systems that may have a large number of robots.
Third, to maximize the utility of a multi-robot system, it needs to support a wide range of tasks. 
In addition, some studies proposed frameworks for automated software development such as MetaGPT \cite{hongmetagpt}, ChatDev \cite{qian2023communicative}, and \cite{dong2024self}. Although broadly relevant, these frameworks are not specifically designed for multi-robot systems.

Recently, a number of studies explored the use of LLMs for multi-robot systems, but their applicability to general-purpose and real-world multi-robot systems still faces significant hurdles. Of particular relevance is LLM2Swarm \cite{StrDorFri2024:neuripsworkshop}, which takes user instructions as input and outputs control policies for  individual robots. Although LLM2Swarm is intended to be task-agnostic, its generality is yet to be experimentally verified. Moreover, LLM2Swarm depends on manually-written demonstration examples, restricting its zero-shot capabilities. 
Other methods such as SmartLLM \cite{kannan2024smart} focus on high-level symbolic planning and do not generate executable low-level control policies.
Furthermore, many methods are tailored for specific tasks—such as formation control \cite{venkatesh2024zerocap, lykov2024flockgpt}, cooperative navigation \cite{yu2023co}, dancing \cite{jiao2023swarm, vyas2024swarmgpt}, or manipulation \cite{mandi2024roco}—and thus lack the generality to address multiple multi-robot tasks. Moreover, the validation in most of the aforementioned methods is performed in simulation, leaving the significant challenge of automated policy deployment on physical multi-robot systems largely unexplored.

Here, we propose \emph{GenSwarm}, an end-to-end system that can automatically generate and deploy multi-robot policies on real-world platforms from natural language instructions for versatile multi-robot tasks. GenSwarm enables users to program a group of robots using simple natural language instructions.
The user instructions are automatically processed via a pipeline of components, including constraint analysis, policy design, policy generation, policy deployment in simulation environments, policy deployment on real-world robots, and policy improvement based on feedback. These components are respectively empowered by LLM agents. 
GenSwarm can automatically deploy the generated code policies as well as the required runtime environments on real-world robots, thus achieving true end-to-end functionality. The automatic deployment is realized by a scalable multi-robot platform that features novel software and hardware architectures.
GenSwarm enables zero-shot policy generation without the need for context learning based on demonstrative examples. When altered or unseen tasks arise, the system can re-generate and re-deploy policies in response to user requests, thereby offering high adaptability for dynamic tasks. Furthermore, due to the use of code policies, the approach is suitable for real-time execution on robots with limited onboard resources.

Extensive experiments demonstrate the high success rate of GenSwarm across various multi-robot tasks. GenSwarm consistently outperforms the
state-of-the-art methods including MetaGPT \cite{hongmetagpt}, CaP \cite{liang2023code}, and LLM2Swarm~\cite{StrDorFri2024:neuripsworkshop}, achieving significant improvements of 37\%, 34\%, and 34\% in average success rate.
GenSwarm provides a promising new paradigm for developing multi-robot systems. Its significance lies in overcoming two limitations of existing work. First, developing multi-robot systems is time-consuming and labor-intensive, and this problem worsens as the number of robots increases. Second, current multi-robot systems lack generality and flexibility. They are often limited to specific tasks or cannot adapt to changing goals and new situations in a timely manner. GenSwarm overcomes these limitations and has the potential to transform the development paradigm of multi-robot systems.

\section*{Results}

\subsection*{Overview of GenSwarm}

\begin{figure}[!b]
    \centering
    \includegraphics[width=\linewidth]{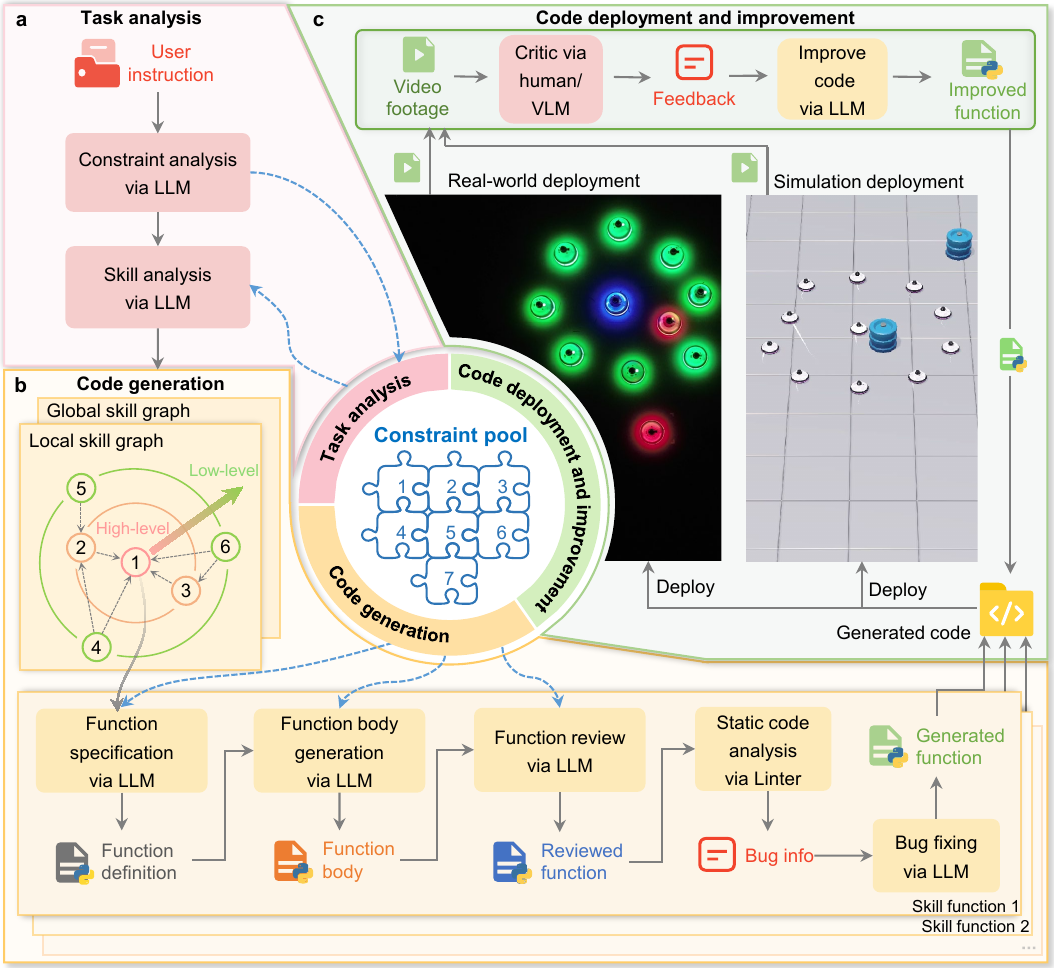}
    \caption{\textbf{The pipeline of GenSwarm.} GenSwarm consists of three modules: task analysis, code generation, and code deployment and improvement. The task analysis module extracts constraints from user instructions and builds a skill library. The code generation module uses a skill graph to hierarchically create and refine Python functions, ensuring constraint alignment and code reusability. Finally, the code deployment and improvement module enables automatic code deployment in simulation and real-world platforms, incorporating feedback from video analysis and human input to refine policies.}
    \label{fig_pipeline}
\end{figure}

The pipeline of GenSwarm consists of three modules: task analysis, code generation, and code deployment and improvement (Fig.~\ref{fig_pipeline}).

The task analysis module takes as input user instructions in the form of natural language about the desired multi-robot task (Fig.~\ref{fig_pipeline}a). For example, to achieve a predator-prey encircling task, the user instruction could be ``{\fontfamily{cmtt}\selectfont\small The robots need to surround the target prey by evenly distributing themselves along a circle with a radius of 1, centered on the prey.}'' 
From the user instruction, an LLM agent extracts constraints that compose a constraint pool. Each constraint specifies what a robot shall or shall not do, such as reaching a target location or not colliding with obstacles. Since the constraint pool comprehensively captures the task requirements, every subsequent step must align with the constraints, thereby ensuring the task is achieved as intended. 
Based on the constraints, an LLM agent generates a skill library where each skill corresponds to a Python function. At this stage, merely the function's name and description are generated; the main body of the function will be generated at a later stage. Skills can be classified as either global or local. Global skills involve global coordination such as goal assignment, whereas local skills are executed onboard each robot based on locally available information.

The code generation module generates the code for the main body of each skill function (Fig.~\ref{fig_pipeline}b). First, a skill graph is constructed by an LLM agent to describe the hierarchical dependencies between the skills and to indicate the constraints that each skill must satisfy. The skill graph guides the code generation process: low-level skills are generated first, and high-level skills thereafter, thereby enhancing code reuse and reducing the need for repetitive code modifications due to lower-level errors.
Once the main body of each skill function has been generated, an LLM agent reviews whether the function aligns with the associated constraints, and makes modifications if necessary. After the review, static code checks are performed, and an LLM agent makes modifications if necessary, ensuring the code is executable.

The code deployment and improvement module realizes automatic code deployment in simulated and real-world robotic platforms (Fig.~\ref{fig_pipeline}c). It relies on novel hardware and software systems, which will be detailed in the following section. It introduces multi-modal feedback mechanisms that can automatically identify issues during execution and effectively adjust policies based on feedback. Specifically, execution results in the simulation can be automatically collected in the format of video clips. A VLM agent assesses the video clips to generate feedback on whether the desired task is successfully completed. In addition, an interface for human feedback is incorporated. It enables users to efficiently modify the policy by providing natural language feedback. 

The global-local control structure can be automatically determined and implemented by the proposed pipeline. During the task analysis stage, the LLM agent judges whether the task requires global skills for global centralized coordination or merely local skills for local distributed execution. This structural decision is encoded in the skill graph to guide code generation, and the resulting architecture determines the deployment model: if global skills are generated, they execute once on the control station using global information, after which local skills are deployed to each robot for distributed execution based only on the local information exposed by the system APIs.

\subsection*{Software and Hardware Platform}

\begin{figure}[!b]
    \centering
    \includegraphics[width=\linewidth]{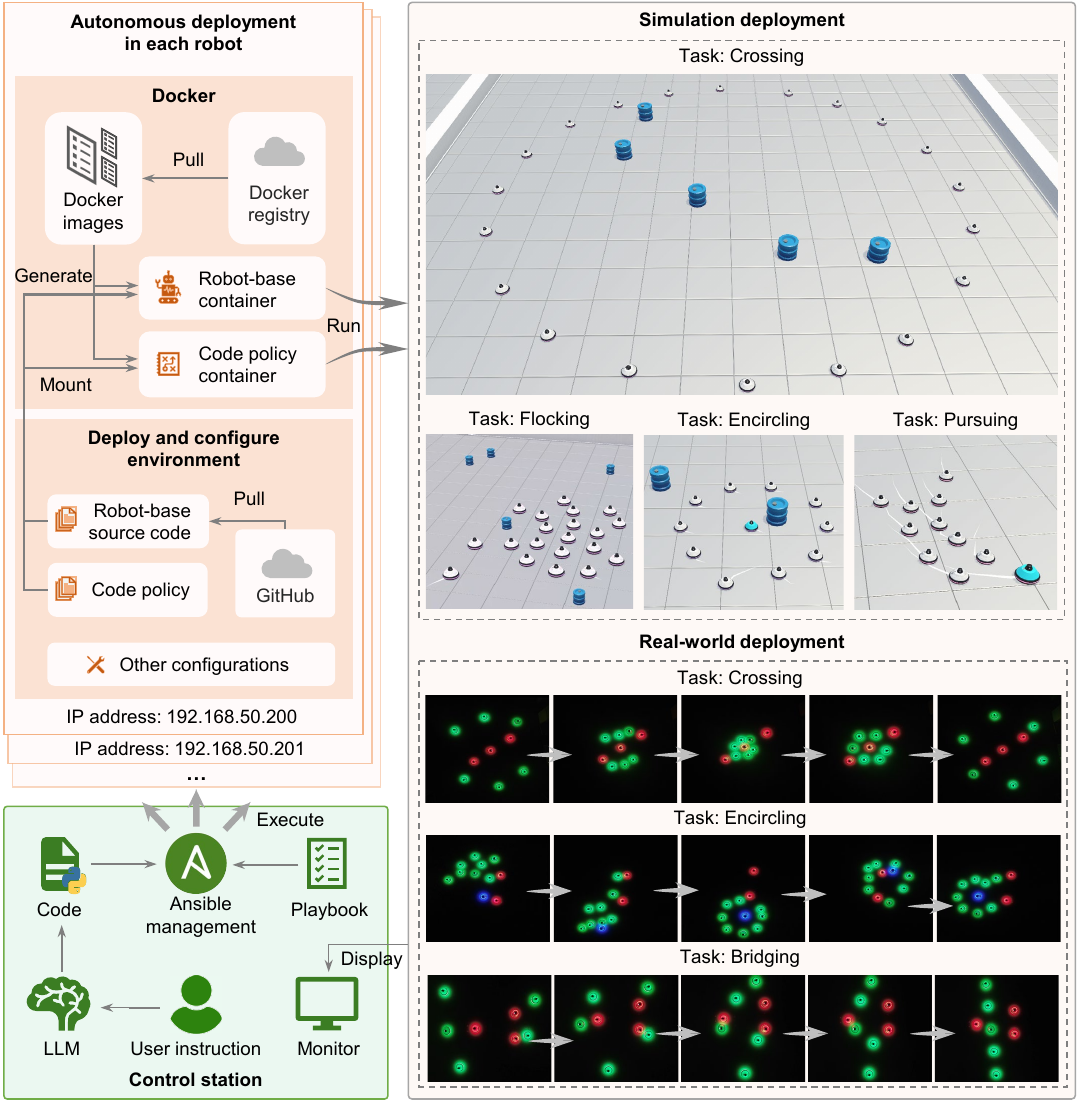}
    \caption{\textbf{Software components of GenSwarm.} 
    A control station generates the required code based on the proposed pipeline and uses Ansible to wirelessly connect to each robot. First, each robot runs Playbook-defined tasks, such as installing and configuring the Docker environment. Then, two pre-built Docker images are pulled: one with the ROS environment for robot operation, and the other with the Python environment for code execution. Once the environments are ready, the generated code is transmitted to all robots and then executed onboard.}
    \label{fig_software}
\end{figure}

\begin{figure}[!b]
    \centering
    \includegraphics[width=\linewidth]{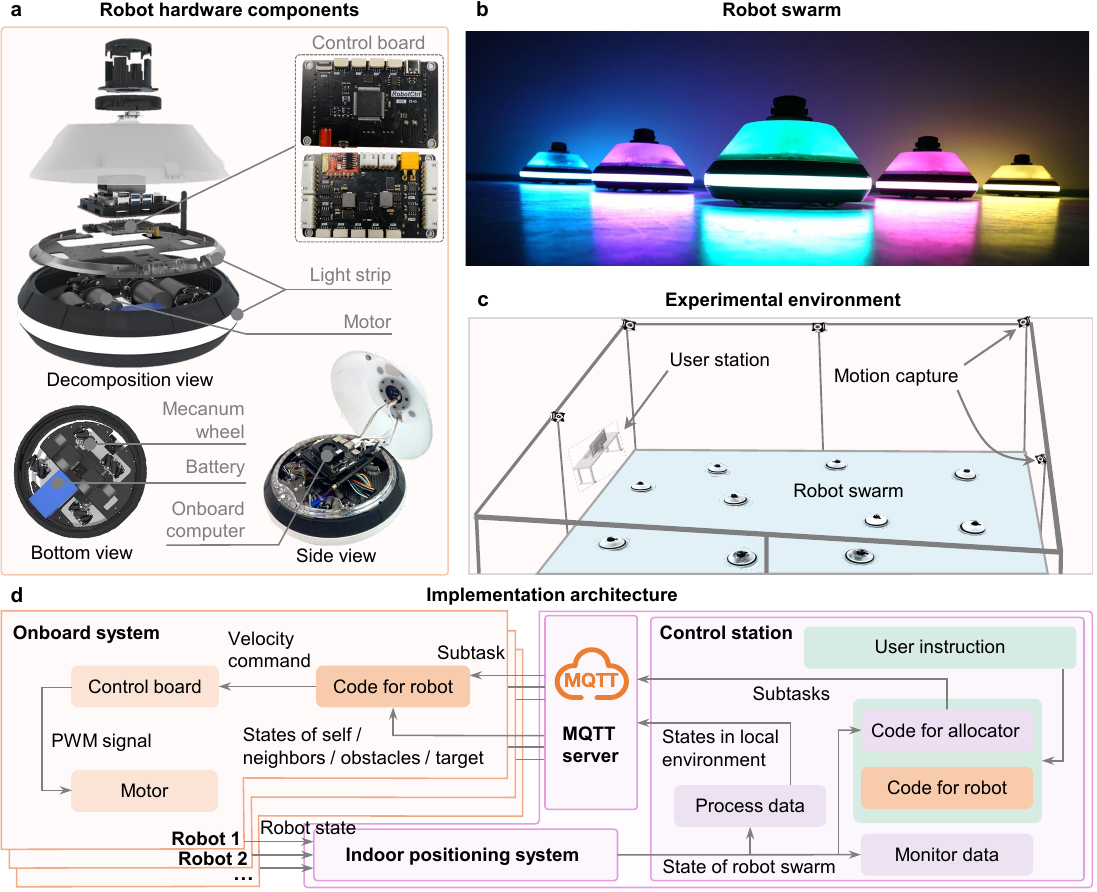}
    \caption{\textbf{Hardware components of GenSwarm.} As a major upgrade of our previous robotic platform \cite{sun2023mean}, each robot has the onboard computational, control, and communication resources to support autonomous code deployment and execution.  The multi-robot system features one-click all start, one-click all sleep, and wireless data retrieval functions that can significantly reduce experimental costs. Since the robots do not have onboard vision systems, the perception was emulated with relevant motion information being collected by an indoor motion capture system, and then distributed to the robots through an MQTT coordination server, ensuring each robot receives only the local information of its surroundings.}
    \label{fig_hardware}
\end{figure}

Automatic deployment is nontrivial as code execution depends on complex runtime environments consisting of various software packages. Manually installing and configuring the runtime environments on each robot would be inefficient as the time required scales linearly with the number of robots. This would make deployment on large-scale multi-robot systems impractical.

GenSwarm possesses a software framework that can automatically deploy both the \textit{generated code} and the \textit{runtime environments} across all the robots in near constant time regardless of the number of robots. In our experiments, automatically deploying the runtime environments on all the robots takes about two minutes, whereas automatically deploying the generated code takes mere seconds. This makes the system particularly well-suited for large-scale multi-robot systems, where consistent and rapid deployment is essential.

The software framework is illustrated in Fig.~\ref{fig_software}. A control station first generates the required code based on the pipeline described earlier and connects with each robot through Ansible via WiFi and SSH (Methods). With predefined automated scripts in the format of Playbook, each robot performs a series of tasks such as installing and configuring the Docker environment. After the Docker environment is ready, two pre-built Docker images are pulled: one containing the ROS environment used for robot operation, and the other containing the Python environment required for code execution. Once the execution environments are ready, the generated code is transmitted to all robots and then executed onboard.
The proposed software framework heavily relies on two techniques, Ansible and Docker (Methods), which work together to simplify and streamline the code deployment on multiple robots. This integration ensures that the deployment process is both repeatable and efficient, drastically reducing the time required to make a group of robots operational. 
Moreover, the framework is designed to be portable across different hardware platforms, a feature enabled by its modular software architecture, which is detailed in the Methods section.

The hardware framework is illustrated in Fig.~\ref{fig_hardware}. 
A new multi-robot platform, which is a major upgrade of our previous robotic platform \cite{sun2023mean}, was developed to support GenSwarm. Each ground robot has onboard computational, control, and communication resources that are necessary for autonomous code deployment and execution \cite{ma2024omnibot}.
Considering that multi-robot experiments involve a large number of operations, such as starting and shutting down robots, we developed novel features for the multi-robot platform such as one-click all start, one-click all sleep, and wireless data retrieval, significantly reducing experimental costs.
It is worth mentioning that the perception of each robot is emulated. Specifically, the motion information of all the robots is collected by an indoor positioning system and then distributed to all robots through an MQTT coordination server so that each robot receives only information about its surroundings (Fig.~\ref{fig_hardware}). The generated code policies access the required information by calling APIs (application programming interfaces) that enforce hard-coded physical limitations. For instance, the sensing API restricts a robot's perception to a fixed local radius (1~m in our experiments), while the motion API clamps velocity commands to a predefined maximum speed. In the future, the sensing API could be realized by onboard vision systems. As the indoor positioning system can provide high-precision measurements, we also exposed our multi-robot system to different levels of measurement noise in real-world experiments. Although performance gradually deteriorates as the noise level increases, the system is still reasonably stable for low to moderate levels of noise. The real-world noise robustness results are provided in Supplementary Fig.~1.

\subsection*{Demonstration of GenSwarm}

\begin{figure}[!b]
    \centering
        \includegraphics[width=\linewidth]{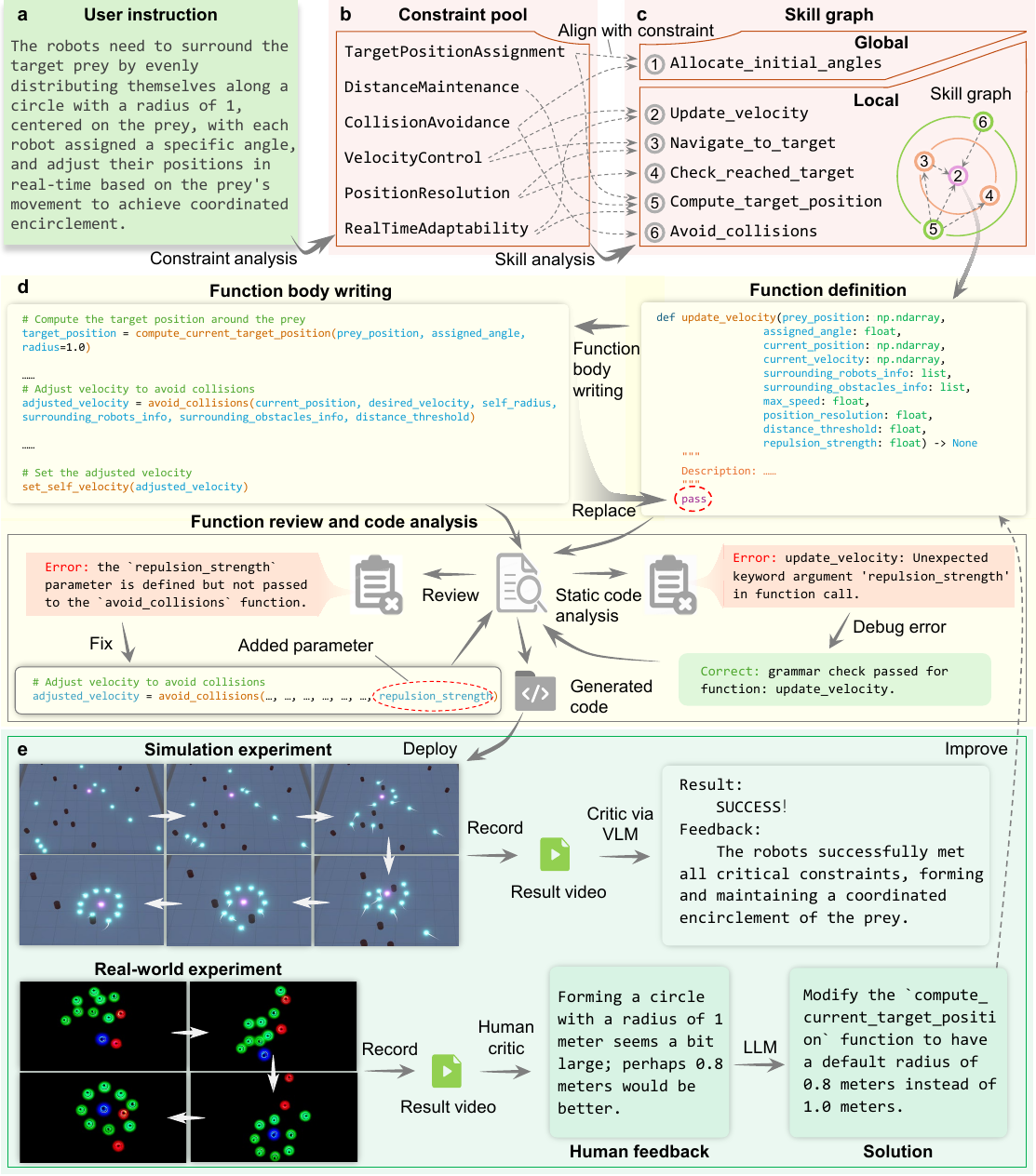}
    \caption{\textbf{A demonstration of the complete workflow of GenSwarm.} \textbf{a.} The user specified a predator-prey encircling task via natural language. \textbf{b.} Six constraints were extracted from the task by LLM agents. \textbf{c.} Six skills were generated and categorized into global (e.g., goal assignment) and local (e.g., update velocity). \textbf{d.} Based on those skills and a consequently generated skill graph, LLM agents generated and reviewed the main-body code of each skill function. \textbf{e.} The code was deployed in simulation environments, reviewed via video feedback by VLM agents, and refined with human feedback. Once validated, it was automatically deployed on real-world robots.}
    \label{fig_demoWorkflow}
\end{figure}

To demonstrate the workflow of GenSwarm, we show the end-to-end generation process of a representative multi-robot task ``predator-prey encircling'' (Fig.~\ref{fig_demoWorkflow}). In this task, multiple predator robots should follow and surround a prey robot that moves randomly. The user instruction is shown in Fig.~\ref{fig_demoWorkflow}a. From the user instruction, six constraints were generated by an LLM agent (Fig.~\ref{fig_demoWorkflow}b). For instance, one of the constraints is ``{\fontfamily{cmtt}\selectfont\small  CollisionAvoidance}'' with the description as ``{\fontfamily{cmtt}\selectfont\small Ensure a minimum distance greater than the sum of the robot's radius, other robots' radii, and a pre-\\defined distance\_threshold from all other robots and obstacles within the perception\\ range}''. 

Based on those constraints, six skills (merely the names and descriptions) are generated (Fig.~\ref{fig_demoWorkflow}c). One of them is a global skill that will be executed on the control station, whereas the others are local skills that will be executed on each robot in a distributed manner. The purpose of the global skill,  named ``{\fontfamily{cmtt}\selectfont\small  Allocate\_initial\_angles}'', is goal assignment, that is, to assign the desired relative angular position of each robot when encircling the target. Goal assignment is a common technique adopted in multi-robot tasks, especially when there is a global constraint such as a geometric shape that multiple robots must satisfy \cite{sun2023mean,zhao2019bearing}.

For tasks like flocking and aggregation that do not involve global goals or constraints, GenSwarm generates distributed policies whose execution merely relies on local information. In contrast, for tasks like shaping that involve global goals or constraints, GenSwarm usually generates combinations of centralized coordination (e.g., position assignment) and distributed control. It is notable that GenSwarm automatically selects and generates control structures, ranging from pure distributed control to hybrid centralized coordination plus distributed control, which reflect the characteristics of the task. This selection process leverages the LLM's strong prior knowledge, learned from its vast training data of robotics literature and code, to associate a high-level task description with a typical and effective control paradigm. In the resulting hybrid architecture, a global skill runs only once on the control station to perform one-time centralized coordination (e.g., initial goal assignment). After this, each robot executes local skills in a distributed manner, relying entirely on local information that is strictly enforced by the system's sensing and motion APIs. The flexibility of augmenting pure distributed control with hybrid centralized coordination allows GenSwarm to adapt across a wide range of multi-robot tasks.

Based on those skills and the consequently generated skill graph that describes their hierarchical dependencies, LLM agents further generate and then review the main-body code of each skill function (Fig.~\ref{fig_demoWorkflow}d). Logical or grammatical code errors can be identified and corrected. Once the skill functions pass the review process and static code analysis, they are automatically deployed and executed in the simulation environment. Then, a VLM agent reviews the video clip of the simulation execution and provides feedback for improving the code (Fig.~\ref{fig_demoWorkflow}e). After that, the generated code is automatically deployed on real-world robotic platforms. It is worth mentioning that human feedback can be incorporated to adjust the code policy (Fig.~\ref{fig_demoWorkflow}e). For instance, if the human feedback is ``{\fontfamily{cmtt}\selectfont\small  Forming a circle with a radius of 1 meter seems a bit large; perhaps 0.8 meters would be better.}'', GenSwarm can adjust the corresponding parameter from 1 to 0.8, enabling efficient human-in-the-loop policy adjustment. 
The ability of human-in-the-loop adjustment provides a practical approach to adapt to newly emerged situations such as robot faults. While this adaptation may not occur in real-time (e.g., at millisecond-level latency), it still offers an effective way for rapid reprogramming and redeployment.

Non-stop one-take videos are attached to show the complete workflows of GenSwarm (Movies~1 and 2). In terms of time consumption, the steps of code generation, deployment onto real-world robots, and improvement based on human feedback took approximately six, two, and two minutes, respectively. The time duration of code generation can be significantly shortened if LLMs' efficiency can be improved in the future. The time of deployment can be shortened to a few seconds if the runtime environment has been pre-installed on the robots and merely the generated code needs to be deployed.
As elaborated above, GenSwarm consists of multiple LLM agents that play different roles. All LLMs and VLMs in GenSwarm are used out-of-the-box without fine-tuning. This design was made to maximize reproducibility, enabling any user to directly deploy the system using off-the-shelf models. Each LLM agent is set up in advance by a prompt involving role description, environment description, robot description, and available APIs. For instance, regarding environment description, the prompt may be ``{\fontfamily{cmtt}\selectfont\small  The environment is composed of a 2D plane with obstacles and robots}''. Regarding robot description, the prompt may be ``{\fontfamily{cmtt}\selectfont\small  The maximum speed of each agent is 0.5~m/s}''. Regarding APIs, the prompt may be ``{\fontfamily{cmtt}\selectfont\small  There are two types of APIs: local and global. Local APIs can only be called by the robot itself, and global APIs can be \\called by a centralized controller}''. Examples of local APIs are ``{\fontfamily{cmtt}\selectfont\small  get\_self\_position}'' and ``{\fontfamily{cmtt}\selectfont\small  get\_surrounding\_robots\_info}''. Examples of global APIs are ``{\fontfamily{cmtt}\selectfont\small  get\_all\_robots\_id}'' and  ``{\fontfamily{cmtt}\selectfont\small  get\_all\\\_robots\_initial\_position}''.
While merely some representative examples are provided here, the complete prompts and APIs can be found in our open-source repository (see Code Availability).

\subsection*{Performance Evaluation}

\begin{figure}[!b]
    \centering
    \includegraphics[width=\linewidth]{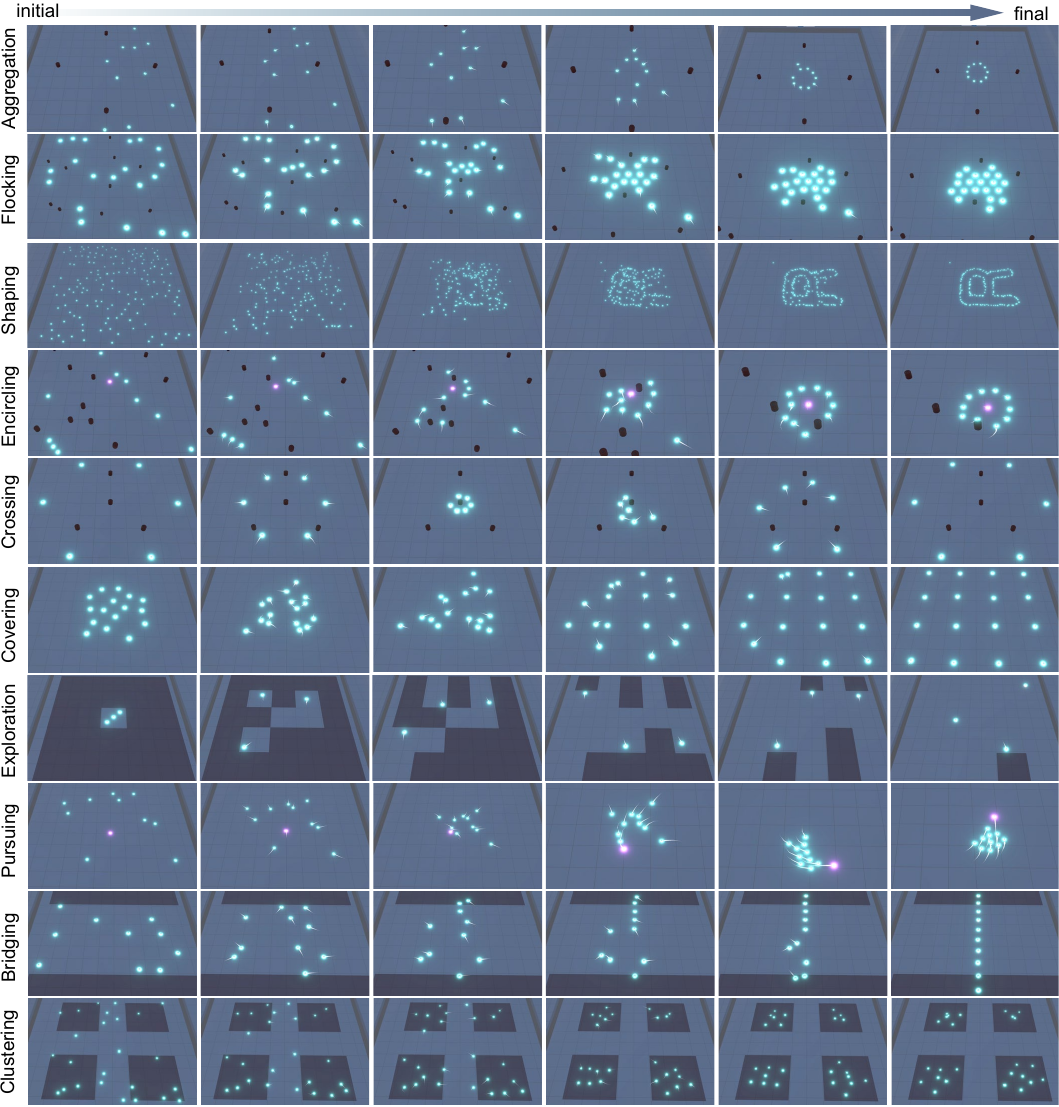}
    \caption{\textbf{Results by GenSwarm for ten multi-robot tasks.} The ten tasks include aggregation, flocking, shaping, encircling, crossing, coverage, exploration, pursuing, bridging, and clustering. These tasks cover a wide range of scenarios, from cooperative to competitive, aiming to comprehensively evaluate the effectiveness of GenSwarm.}
    \label{fig_tenTask}
\end{figure}

\begin{figure}[!b]
    \centering
    \includegraphics[width=\linewidth]{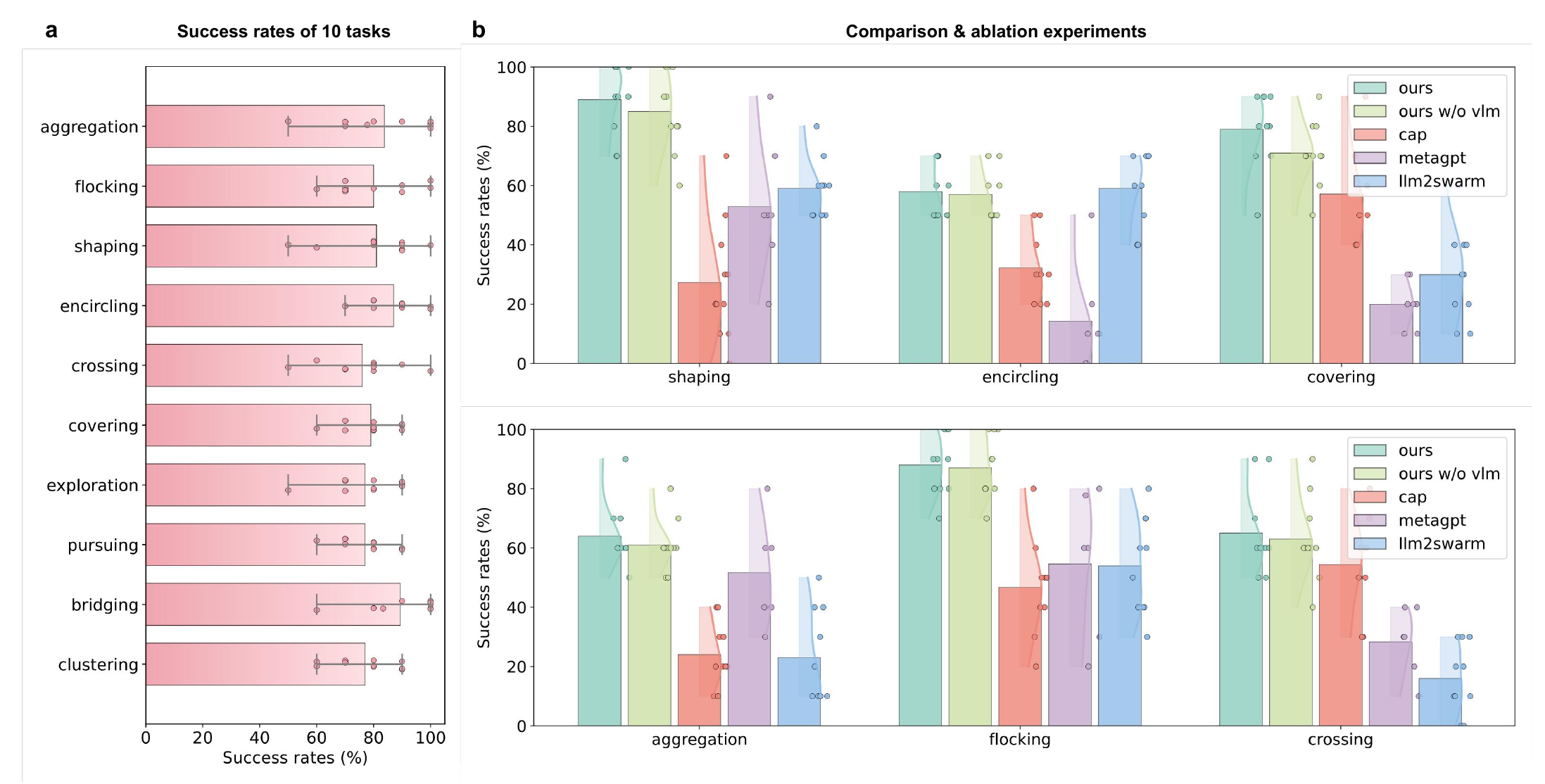}
    \caption{
    \textbf{Success rate of different LLMs on different tasks.} 
    \textbf{a.} The success rates of GenSwarm across ten multi-robot tasks. One hundred independent trials, from user instructions to code execution in simulation, were run for each task. Hence, 1,000 trials in total were run and the average success rate was 81\%.
    \textbf{b.} The comparison between GenSwarm, CaP, MetaGPT, LLM2Swarm and GenSwarm without VLM feedback across six representative tasks. One hundred independent trials, from user instructions to code execution in simulation, were run for each method and each task. The average success rates of GenSwarm, GenSwarm without VLM, CaP, MetaGPT, and LLM2Swarm were 74\%, 71\%, 40\%, 37\%, and 40\%, respectively.
    }
    \label{fig_comparison}
\end{figure}

\textbf{Different tasks: }The performance of GenSwarm was evaluated on ten different multi-robot tasks, including aggregation, flocking, shaping, encircling, crossing, coverage, exploration, pursuing, bridging, and clustering (Fig.~\ref{fig_tenTask}). These tasks cover a wide range of scenarios, from cooperative to competitive, aiming to comprehensively evaluate the effectiveness of GenSwarm.
Details of the tasks and the evaluation metrics are given in Methods.
The LLM used here was o1-mini, one of the state-of-the-art LLMs. One hundred independent trials, starting from user instruction to code execution in simulation, were run for each of the ten tasks. 
The average success rate over the 1,000 trials for 10 tasks was 81\%. 
The respective success rate for each task is presented in Fig.~\ref{fig_comparison}a. 

\textbf{Different methods:} GenSwarm has been compared to three state-of-the-art methods, MetaGPT \cite{hongmetagpt}, CaP \cite{liang2023code}, and LLM2Swarm~\cite{StrDorFri2024:neuripsworkshop}. To ensure a fair comparison, all baseline frameworks were configured according to their native design paradigms (zero-shot or few-shot). For few-shot methods like CaP and LLM2Swarm, we provided high-quality multi-robot examples—handcrafted for CaP and drawn from the official LLM2Swarm repository for the latter—with minimal adaptations for our platform. The complete prompts and code examples used for all baselines are publicly available for reproducibility (see Code Availability).
Moreover, GenSwarm without VLM feedback was also compared. One hundred independent trials, from user instructions to code execution in simulation, were run for each method and each task. Six representative tasks were selected, and hence 2,400 trials in total were run. The LLM used here was GPT-4o.
The comparison results are shown in Fig.~\ref{fig_comparison}b. As can be seen, GenSwarm achieved the highest average success rate, which was 74\%, across different tasks. The average success rates of GenSwarm without VLM, CaP, MetaGPT, and LLM2Swarm were 71\%, 40\%, and 37\%, and 40\%, respectively. GenSwarm consistently outperforms the baselines, achieving 34\%, 37\%, and 34\% higher success rate than LLM2Swarm, MetaGPT, and CaP.
We also compared using finer-grained task-specific metrics, with detailed results presented in Supplementary Fig.~2. It reveals that GenSwarm consistently achieves superior performances across different tasks.

In addition, Supplementary Fig.~2 includes a benchmark against fine-tuned state-of-the-art (SOTA) expert controllers. For distributed tasks such as flocking and aggregation, we used the classic Boids model~\cite{reynolds1987flocks}; for hybrid tasks such as shaping, crossing, covering, and encircling, we combined an optimal assignment algorithm based on the Hungarian method with VR-ORCA~\cite{guo2021vr}. While these SOTA controllers achieve higher average performance, GenSwarm's best-performing policies reach comparable levels in some cases, demonstrating its potential to deliver high-quality solutions without extensive manual tuning.

\textbf{Different LLMs: }
By comparing Figs.~\ref{fig_comparison}a and~\ref{fig_comparison}b, it can be seen that different LLMs (o1-mini and GPT-4o) lead to similar success rates though there are subtle variations. We further expanded the comparison to include two additional prominent LLMs, DeepSeek-V3 and Claude-3.7-Sonnet. Results consistently show high success rates across these models (Supplementary Fig.~3), suggesting general applicability of GenSwarm across different types of LLMs.

\textbf{Different prompts: }The user instructions have a significant impact on the performance of GenSwarm. For instance, comprehensive instructions tend to yield better results, while ambiguous ones may lead to failures (see examples in Supplementary Fig.~4). To systematically analyze this effect, we designed seven representative prompt types that range from unstructured to highly structured formats: 1) {Plain-Compound (Cohesive)}, which integrates both the task objective and policy into a linguistically coherent paragraph; 2) {Plain-Compound}, which strictly concatenates the verbatim text from the objective-only and policy-only prompts; 3) {Plain-Objective}, which provides only the objective but no policy; 4) {Plain-Policy}, which provides only the policy but omits task objectives; 5) {Plain-Narrative}, which uses natural, human-like language to describe the task but lacks formal structure or policy details; 6) {Structured-Objective}, which restructures the instruction into a ``description-goal-constraint'' format; and 7) {Structured-Policy}, which adds explicit policies and constraints on top of the structured prompt.
Examples of the seven prompt types for an encircling task are given in Supplementary Fig.~5. Moreover, all the previous evaluations were conducted using the Plain-Compound (Cohesive) prompt type.

As shown in Supplementary Fig.~6, the inclusion of explicit policy instructions is helpful for achieving high task success rates. Specifically, prompt types that contained policy instructions—
{Plain-Compound (Cohesive)} (78\%), {Plain-Compound} (74\%)
, {Plain-Policy} (74\%),  and {Structured-Policy} (74\%)—yielded higher success rates. Conversely, prompts lacking this information, such as {Plain-Objective} (56\%), {Plain-Narrative} (57\%), and {Structured-Objective} (57\%), resulted in significantly lower success rates. This demonstrates that the presence of policy instructions is more impactful than the prompt's format (i.e., natural vs. structured language). Among the top performers, {Plain-Policy} emerges as a particularly practical choice, leading to high success rates while offering the simplicity of concise natural language inputs.

\section*{Discussion}

This work introduced GenSwarm, an end-to-end system that automatically generates and deploys code policies for versatile multi-robot tasks. 
As a significant step toward end-to-end generation, GenSwarm presents a novel paradigm that could potentially disrupt the current development process of multi-robot systems. However, GenSwarm has some limitations that could be addressed in the future. First, this study focussed on decision-making and control. Aspects such as sensing and navigation, which are important for practical applications, have not been incorporated. Developing and integrating onboard sensing into the system would be a valuable direction for future research. Second, we focus on the framework’s generality and end-to-end automation in this work, rather than the novelty or optimality of the generated policies or collective behaviors.
Generating more sophisticated or optimal policies is an important future research topic, which might be challenging to achieve when relying solely on LLMs. Combining language models with other techniques, such as multi-agent reinforcement learning, could be a promising approach. Compared to language models, reinforcement learning is better suited for generating more sophisticated policies, making it a valuable complement to GenSwarm. 
Third, GenSwarm generates policies from scratch rather than re-using existing ones. This design choice was made to achieve zero-shot capability. Nevertheless, re-using a behavioral repertoire of previously generated solutions is a valuable direction for future research.

\section*{Methods}

\subsection*{Ten Multi-Robot Tasks}

The ten multi-robot tasks considered in this work are aggregation, flocking, shaping, encircling, crossing, coverage, exploration, pursuing, bridging, and clustering. The following gives the user instructions and evaluation metrics of each task. Multiple metrics may be used to evaluate a task from different aspects. It should be noted that these metrics are used solely for post-evaluation but not incorporated into the policy generation pipeline.
A task is regarded as successful when all of its corresponding metrics exceed certain predefined thresholds. In this way, we can automatically calculate the success rate of each task. The termination of a simulation trial is triggered when the execution time exceeds certain values or the task has finished in the sense that, for example, all the robots succeed in reaching their desired positions.

\medskip
\noindent\textbf{Aggregation task}: \emph{User instruction:} ``{\fontfamily{cmtt}\selectfont\small {The robots need to aggregate as quickly as possible and avoid colliding with each other.}}''

\emph{Evaluation metric:}  
Maximum of minimum distances, denoted as $d_{\text{maxmin}}$: It quantifies the largest minimum distance between each robot and its closest neighbor. It is defined as
\begin{equation}
d_{\text{maxmin}} = \max_i \min_{j \neq i} ||\mathbf{p}_i - \mathbf{p}_j||
\end{equation}
where $||\mathbf{p}_i - \mathbf{p}_j||$ is the Euclidean distance between robots $i$ and $j$. 
The task is regarded as successful if the value of this metric is less than 1.

\medskip
\noindent\textbf{Flocking task}: \emph{User instruction:} 
``{\fontfamily{cmtt}\selectfont\small  The robots must form a cohesive flock, cooperating with all others in the environment. The three main behaviors are cohesion, alignment, and separation: cohesion maintains connectivity, alignment ensures synchronized movement, and separation prevents collisions by keeping robots at least 0.5 meters apart.}''

\emph{Evaluation metrics:} 
The flocking task is evaluated based on two metrics. The task is treated as successful when both metrics exceed their corresponding thresholds.

1) Spatial Variance, denoted as $\text{Var}_{\text{spat}}$: It quantifies how spread out the robots are. It is defined as
\begin{equation}
\text{Var}_{\text{spat}} = \sum_{d \in \{x, y\}} \text{Var}(P_d)
\end{equation}
where $\text{Var}(P_d)$ is the variance of the robot positions along the $d$ dimension (either $x$ or $y$). 
The task is regarded as successful if the value of this metric is less than 1.

2) Mean Dynamic Time Warping (DTW) Distance, denoted as \( d_{\text{DTW}} \): This metric quantifies the similarity between the trajectories of all robots. It is defined as
\begin{equation}
d_{\text{DTW}} = \frac{1}{M} \sum_{i < j} \text{DTW}(\mathbf{T}_i, \mathbf{T}_j)
\end{equation}
where \( M \) is the total number of robot pairs, \( \mathbf{T}_i \) is the trajectory of robot \( i \), and \(\text{DTW}(\mathbf{T}_i, \mathbf{T}_j)\) is the DTW distance between \(\mathbf{T}_i\) and \(\mathbf{T}_j\).
Here, \(\mathbf{T}_i = \{(x_i^1, y_i^1), \dots, (x_i^{1,000}, y_i^{1,000})\}\) and \(\mathbf{T}_j = \{(x_j^1, y_j^1), \dots, (x_j^{1,000}, y_j^{1,000})\}\). The DTW distance between them is defined as \cite{Toohey2015}
\begin{equation}
\text{DTW}(\mathbf{T}_i, \mathbf{T}_j) = \min_{W \in \Omega} \sum_{(a,b)\in W} d\bigl((x_i^a, y_i^a), (x_j^b, y_j^b)\bigr),
\end{equation}
where \( W \) is the warping path, a valid alignment between \(\mathbf{T}_i\) and \(\mathbf{T}_j\) that satisfies constraints such as boundary, continuity, and monotonicity. The function \( d(\cdot,\cdot) \) is the Euclidean distance.

The task is regarded as successful if the value of this metric is less than 500. Since each trajectory has 1,000 points, the threshold of 500 indicates that the average distance between pairs of points across two trajectories is less than 0.5. 

\medskip
\noindent\textbf{Shaping task}: \emph{User instruction:}  
``{\fontfamily{cmtt}\selectfont\small The robots need to form a specific shape, with each robot assigned a unique point on that shape. The task requires each robot to move towards and maintain its assigned position on the target shape.}''

\emph{Evaluation metric:} 
Procrustes Distance, denoted as $d_{\text{proc}}$: It quantifies the similarity between the robot positions and the target shape. It is defined as
\begin{equation}
d_{\text{proc}} = \min_{\mathbf{Q}} \frac{1}{N}\sum_{i=1}^{N} ||\mathbf{p}_{i} - \mathbf{Q} \mathbf{p}_{i,\text{T}}||^2
\end{equation}
where $N$ is the total number of robots, $\mathbf{p}_{i}$ is the current position of robot $i$, $\mathbf{p}_{i,\text{T}}$ is the target position for robot $i$ on the straight line, and $\mathbf{Q}$ is the optimal permutation matrix to be solved. The task is regarded as successful if the value of this metric is less than 0.1.

\medskip
\noindent\textbf{Encircling task}: \emph{User instruction:}  
``{\fontfamily{cmtt}\selectfont\small The robots need to surround the target prey by evenly distributing themselves along a circle with a radius of 1, centered on the prey. Each robot is assigned a specific angle, and they must adjust their positions in real-time based on the prey's movement to achieve coordinated encirclement.}''

\emph{Evaluation metric:} 
Mean distance error, denoted as $d_{\text{error}}$: It quantifies the average deviation of the robots' distances from the desired radius. It is defined as
\begin{equation}
d_{\text{error}} = \frac{1}{N} \sum_{i=1}^{N} \big| ||\mathbf{p}_{i} - \mathbf{p}_{\text{prey}}|| - r_{\text{desired}} \big|
\end{equation}
where $N$ is the total number of robots, \(\mathbf{p}_{i}\) is the position of robot \(i\), \(\mathbf{p}_{\text{prey}}\) is the position of the prey, and \(r_{\text{desired}}\) is the desired radius. The task is regarded as successful if the value of this metric is less than 0.1.

\medskip
\noindent\textbf{Crossing task}: \emph{User instruction:}  
``{\fontfamily{cmtt}\selectfont\small Each robot must maintain a distance of at least fifteen centimeters from other robots and obstacles to avoid collisions while moving to the \\target point, which is the position of the robot that was farthest from it at the \\initial moment.}''

\emph{Evaluation metric:} 
Target Reach Ratio, denoted as $\rho _{\text{reach}}$: It quantifies the proportion of robots that successfully reached their target positions within a certain tolerance distance (typically 0.1 meters). It is defined as
\begin{equation}
\rho _{\text{reach}} = \frac{N_{\text{reach}}}{N}
\end{equation}
where $N_{\text{reach}}$ is the number of robots reached targets. 
The task is regarded as successful if the value of this metric is equal to 1.

\medskip
\noindent\textbf{Coverage task}: \emph{User instruction:}  
``{\fontfamily{cmtt}\selectfont\small Divide the environment into sections equal to the number of robots. Each robot needs to move to the center of its assigned section to achieve full coverage of the environment.}''

\emph{Evaluation metrics:} 
The coverage task is evaluated based on two metrics. The task is treated as successful when both metrics exceed their corresponding thresholds.

1) Area Ratio, denoted as $\rho _{\text{area}}$: It quantifies how much of the total area is occupied by the robots. It is defined as
\begin{equation}
\rho _{\text{area}} = \frac{A_{\text{occupied}}}{A_{\text{total}}}
\end{equation}
where $A_{\text{occupied}}$ is the area occupied by the robots and $A_{\text{total}}$ is the total available area. More specifically, $A_{\text{occupied}}$ is calculated as $A_{\text{occupied}}=(x_{\max} - x_{\min}) \times (y_{\max} - y_{\min})$, where $x_{\max}$ and $x_{\min}$ are the maximum and minimum x-coordinates among all the robots, respectively. The task is regarded as successful if the value of this metric is greater than 0.8.

2) Variance of Nearest Neighbor Distances, denoted as $\text{Var}_{\text{NND}}$: It quantifies how evenly spaced the robots are from their nearest neighbors. It is defined as
\begin{equation}
\text{Var}_{\text{NND}} = \text{Var}(d_{\text{near}})
\end{equation}
where $d_{\text{near}}$ is the Euclidean distance between each robot and its nearest neighbor. 
The task is regarded as successful if the value of this metric is less than 0.1.

\medskip
\noindent\textbf{Exploration task}: \emph{User instruction:}  
``{\fontfamily{cmtt}\selectfont\small The robots need to explore all the unknown areas. \\You are required to assign an optimal sequence of exploration areas to each robot \\based on the number of robots and the unexplored regions, and then the robots will \\gradually explore these areas.}''

\emph{Evaluation metric:} 
Landmark Visit Ratio, denoted as $\rho _{\text{visit}}$: It quantifies the proportion of unexplored areas (landmarks) that were successfully visited by the robots. It is defined as
\begin{equation}
\rho _{\text{visit}} = \frac{N_{\text{visit}}}{N_{\text{total}}}
\end{equation}
where $N_{\text{visit}}$ is the number of visited landmarks and $N_{\text{total}}$ is the total number of landmarks. A landmark is considered visited if a robot comes within a certain distance (e.g., 0.1 meters) of the landmark center. 
The task is regarded as successful if the value of this metric is equal to 1.

\medskip
\noindent\textbf{Pursuing task}: \emph{User instruction:}  
``{\fontfamily{cmtt}\selectfont\small Engage in flocking behavior with all robots on the map, moving toward the lead robot. The lead robot's movement is unpredictable, so maintain cohesion by staying connected, ensure alignment by moving in sync, and uphold separa-\\tion by keeping a safe personal space. Additionally, be cautious to avoid collisions with any obstacles in the environment.}''

\emph{Evaluation metrics:} 
The pursuing task is evaluated based on two metrics. The task is treated as successful when both metrics exceed their corresponding thresholds.

1) Average distance to prey, denoted as $d_{\text{avg-prey}}$: It measures the average distance between all robots and the prey's position. It is defined as:
\begin{equation}
d_{\text{avg-prey}} = ||\mathbf{p}_{\text{avg}} - \mathbf{p}_{\text{prey}}||
\end{equation}
where \(\mathbf{p}_{\text{avg}}\) is the average position of all robots, and \(\mathbf{p}_{\text{prey}}\) is the prey's position. The task is regarded as successful if the value of this metric is less than 1.

2) Maximum of minimum distances, denoted as $d_{\text{maxmin}}$: It quantifies the largest minimum distance between each robot and its closest neighbor. 
Its definition is equivalent to the metric used in the aggregation task. The task is regarded as successful if the value of this metric is less than 1.

\medskip
\noindent\textbf{Bridging task}: \emph{User instruction:} 
``{\fontfamily{cmtt}\selectfont\small The robots need to evenly form a straight line bridge at the position where x is equal to zero within the range of y between minus two and two.}''

\emph{Evaluation metric:} 
Procrustes distance, denoted as $d_{\text{proc}}$: It quantifies the shape similarity between the robots' final positions and the target straight line. Its definition is the same as the metric used in the shaping task and hence omitted here. The task is regarded as successful if the value of this metric is less than 0.1.

\medskip
\noindent\textbf{Clustering task}: \emph{User instruction:}  
``{\fontfamily{cmtt}\selectfont\small Robots with initial positions in the same quadrant need to cluster in the designated area of that corresponding quadrant.}''

\emph{Evaluation metric:}
Achievement Ratio, denoted as \( r_{\text{achieve}} \): This metric evaluates the proportion of robots that successfully reach their assigned target regions based on their initial quadrant classification. It is defined as
\begin{equation}
r_{\text{achieve}} = \frac{\sum_{q=1}^{4} N_{q,\text{achieved}}}{N_{\text{total}}}
\end{equation}
where \( N_{q,\text{achieved}} \) represents the number of robots in quadrant \( q \) that reach the corresponding target region within a tolerance of \( 0.1 \), and \( N_{\text{total}} \) is the total number of robots. 
The task is considered successful if \( r_{\text{achieve}} = 1 \), indicating all robots meet the criteria.

\subsection*{Details of Software Architecture}

We designed a modular architecture consisting of seven core modules, each containing multiple classes (Supplementary Fig.~7). These classes have inheritance, association, and composition relationships, which enhance system design by enabling code reuse, modularity, and flexibility. The Core Module defines the interfaces between modules, ensuring that they can seamlessly integrate into the system as long as they follow these interfaces. The Skill Module handles skill library operations, including the representation of skills as a skill graph and the functionality to construct, modify, and extend this graph. The Action Module contains all action nodes that encompass tasks such as analyzing constraints, designing functions, writing code, performing syntax checks, and debugging, all guided by interactions with the LLM. The Environment Module encompasses various simulation environments or real-world scenarios, the Constraint Module handles constraint-related tasks, the File Module manages file storage, and the Feedback Module processes all feedback.

The core of the architecture is the Core Module, which includes a set of interfaces and base classes that provide shared interfaces and core functionality to the system’s other modules. Specifically, the Core Module uses BaseActionNode, ActionNode, and CompositeActionNode to implement the Composite Pattern\cite{gamma1994patternBook}, ensuring consistent usage of single and composite action nodes, thereby effectively simplifying the system’s complexity. All actions in the Action Module inherit directly from ActionNode, and these action nodes form the core functionality required by the framework. Take GenerateFunctions, a CompositeActionNode, as an example: it consists of four actions—DesignFunctionAsync, WriteFunctionsAsync, GrammarCheckAsync, CodeReviewAsync, and WriteRun—executed in a specific order. This composite node can be reused whenever GenerateFunctions is needed, eliminating the need to rebuild the sequence. Furthermore, GenerateFunctions itself can be treated as a standard ActionNode, maintaining consistency in how single and composite actions are handled.

Moreover, the Core Module provides several key interfaces to support the system's diverse requirements. The Feedback interface provides a unified handling mechanism for HumanFeedback, CodeBug, and CriticFeedback, as shown in the Feedback Module. The BaseFile interface standardizes the handling of various file types, coverage code files, program logs, and Markdown documents, as shown in the File Module. The BaseEnvironment interface offers standardized access points for different simulation environments, allowing the system to easily adapt to various runtime environments, as shown in the Environment Module. The BaseGraphNode interface unifies the operations of ConstraintNode and SkillNode, ensuring consistency between them and simplifying the establishment of mapping relationships between the two. SkillNode forms SkillLayer, and multiple SkillLayers can form a SkillGraph, constituting the layered structure of the framework mentioned above, as shown in the Skill Module.

The proposed software architecture has the following features. First, by defining generic interfaces and base classes, it achieves a high degree of scalability, allowing the system to easily introduce new functional modules or replace existing ones while maintaining overall system stability. Second, by leveraging the composite pattern technique, which organizes objects into tree-like structures, it unifies the handling of individual and composite skills. Individual skills serve as leaf nodes, while composite skills are represented as branches, allowing users to easily build complex skill structures by combining and nesting different skill nodes. Third, the system supports both simulation and real-world experimental platforms, achieving a unified access point across different platforms. 

\subsection*{Details of Automatic Deployment}

The following introduces the tools of Ansible and Docker and how they are integrated into our automatic deployment framework.

Ansible is an open-source automation tool that allows tasks to be performed consistently across multiple devices. In our framework, it is used to establish wireless connections with robots via SSH (Secure Shell, enabling secure remote communication) and execute predefined playbooks—scripts that outline the steps for deployment. For example, Ansible ensures directories are created, source code is copied, dependencies are installed, and permissions are set on all robots simultaneously. This consistency reduces human error and eliminates the need for manual intervention on individual robots.

The Docker environment includes all the necessary components for seamless robot operation and code execution. It is equipped with ROS (Robot Operating System), a middleware essential for controlling and managing robotic systems. Additionally, it includes a Python runtime preconfigured with all dependencies required to execute the LLM-generated code. 

The deployment process begins with Ansible transferring to each robot the necessary files, such as Python scripts, ROS configuration files, and Dockerfiles, which define the instructions to build the containerized environment for running the code. Once these files are in place, Ansible uses Docker to build the Docker image, packaging the runtime environment and all necessary dependencies. It then pulls and tags prebuilt images to reduce setup time by downloading existing configurations. Afterward, Ansible launches the containers, starting the robot-specific workspace and preparing it for code execution. Inside the container, the code is compiled to ensure compatibility with the ROS environment. Finally, the LLM-generated code is executed via ROS launch files, allowing the experiment to run automatically without further manual intervention.

\section*{{Data availability}}

 The data in this study are available in the main text and the supplementary information. Other source data are available from the corresponding author upon reasonable request.

\section*{Code availability}

The code of the proposed GenSwarm system is available online: \url{https://github.com/WindyLab/GenSwarm}.

\bibliography{references}
\bibliographystyle{spmpsci}

\section*{Acknowledgments}

The authors would like to thank Jialing Lyu for her help in editing the videos. This work was partially supported by the STI 2030-Major Projects (Grant No. 2022ZD0208800) and National Natural Science Foundation of China (Grant No. 62473320, 62473017). Roderich Gro{\ss} acknowledges support by the OpenSwarm project, which has received funding from the European Union’s Horizon Europe Framework Program under Grant Agreement No. 101093046 and by Robotics Institute Germany (BMBF Grant No. 16ME1001).

\section*{Author contributions}

S.Z., R.G., R.Z., and M.C. designed the research and wrote the paper; 
W.J., H.C., M.C., G.Z., and L.X. performed research and analyzed data.

\section*{Competing interests}

The authors declare no competing financial interests.

\section*{Correspondence}

Correspondence and requests for materials should be addressed to Shiyu Zhao.

\newpage

\section*{Supplementary Information}

This file includes Supplementary Figures 1–8 and two Supplementary Movies demonstrating the experimental results.

\section*{Supplementary Figures}

\setcounter{figure}{0}
\renewcommand{\figurename}{Supplementary Fig.}

\begin{figure}[h]
    \captionsetup{}
    \centering
    \includegraphics[width=1\linewidth]{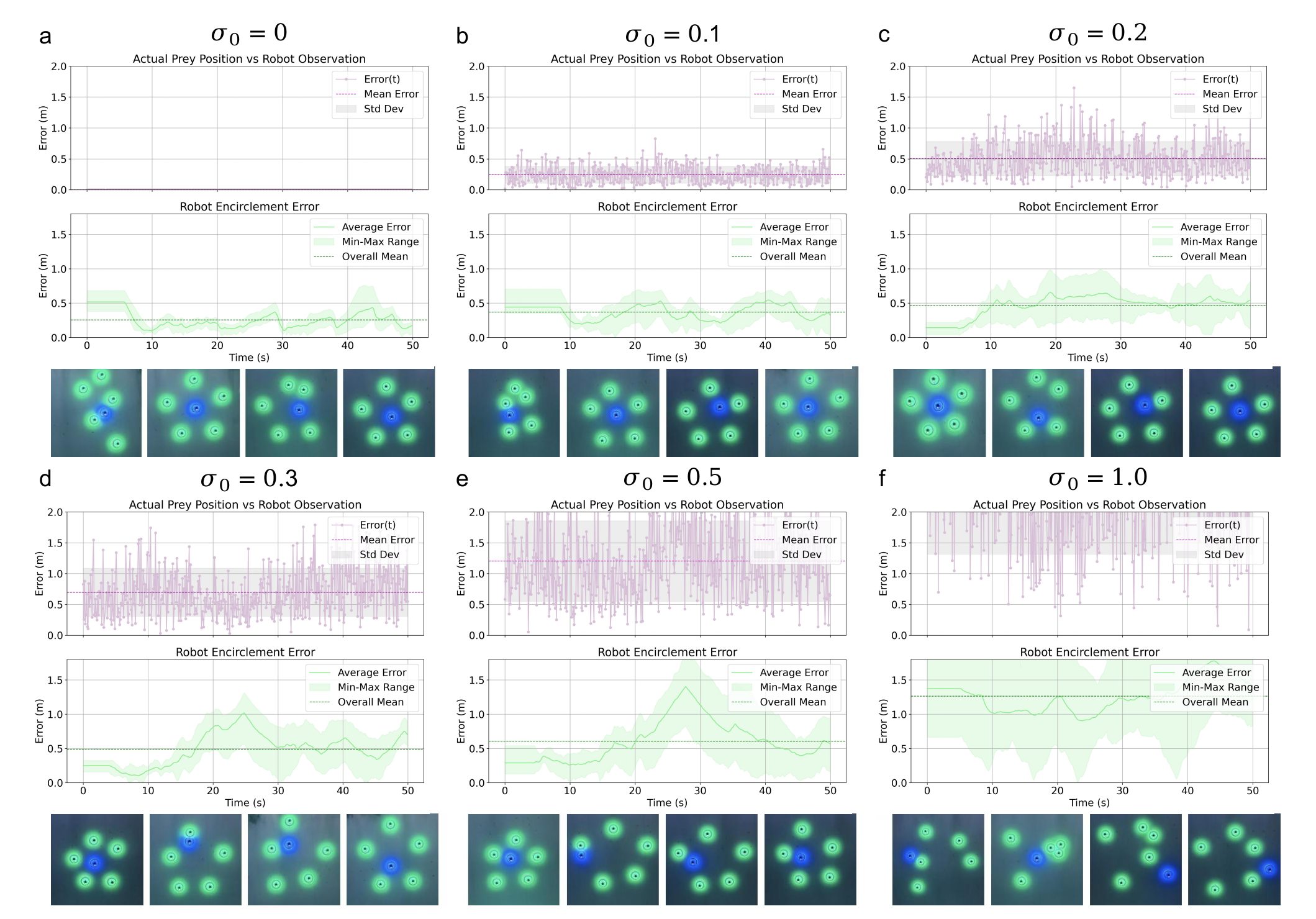}
    \caption{
    \textbf{Performance evaluation for the encircling task subject to sensor noises.}
    Encircling task performance under different noise levels: as noise increases from $\sigma_0 = 0$ to $1.0$, tracking error generally increases.}
    \label{noise}
\end{figure}

\newpage

\begin{figure}[p]
    \centering
    \captionsetup{}
    \includegraphics[width=1\linewidth]
    {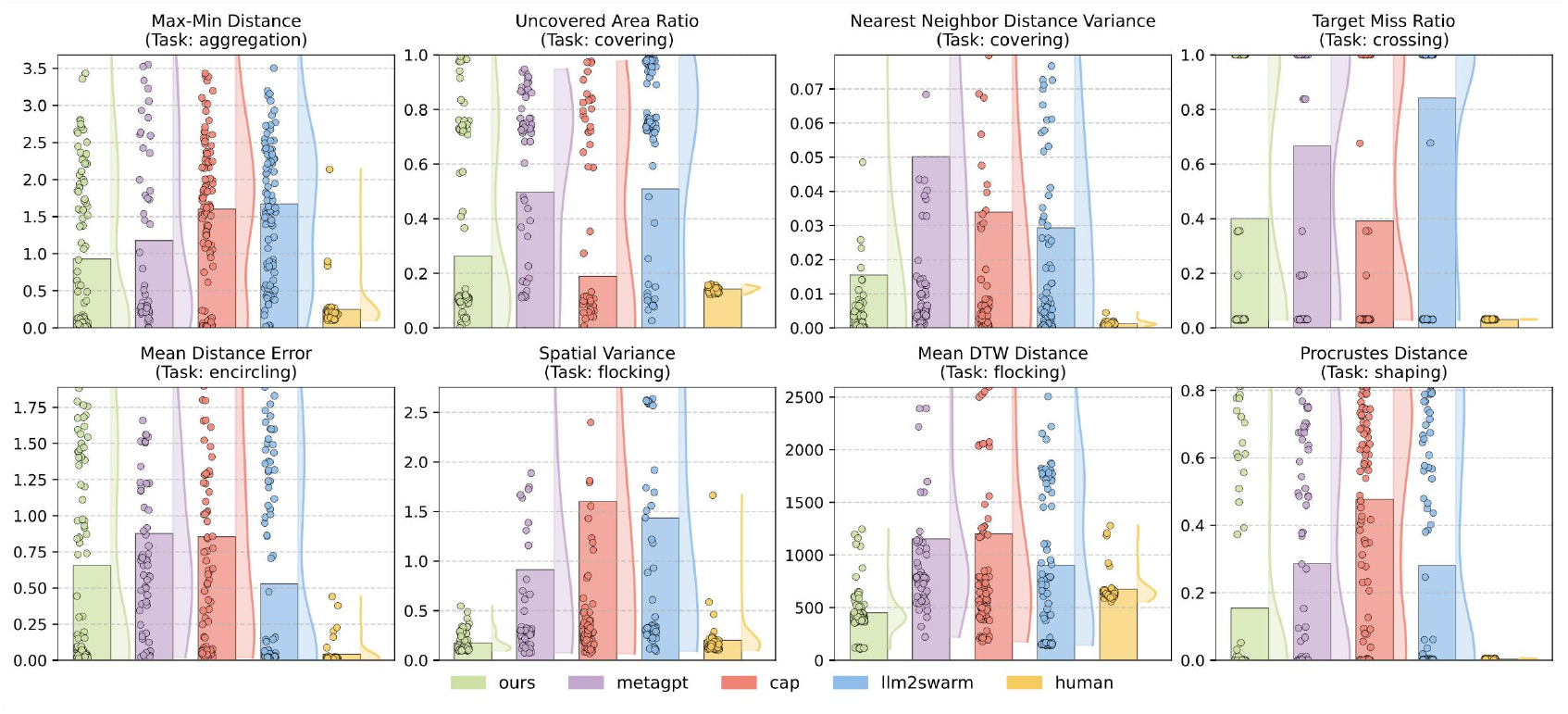}
    \caption{
        \textbf{Performance comparison between different methods.} The figure compares our method (GenSwarm) against three baselines (MetaGPT, CaP, LLM2Swarm) and fine-tuned state-of-the-art (SOTA) expert controllers on six tasks over one hundred trials each. All eight metrics are normalized for a lower-is-better evaluation. It can be seen that GenSwarm achieves the best results among the LLM-based methods, and its best-performing policies are competitive with the SOTA controllers.
    }
    \label{performance_ranking_comparison}
\end{figure}

\clearpage
\begin{figure}[p]
    \centering
    \captionsetup{labelfont={}}
    \includegraphics[width=1\linewidth]
    {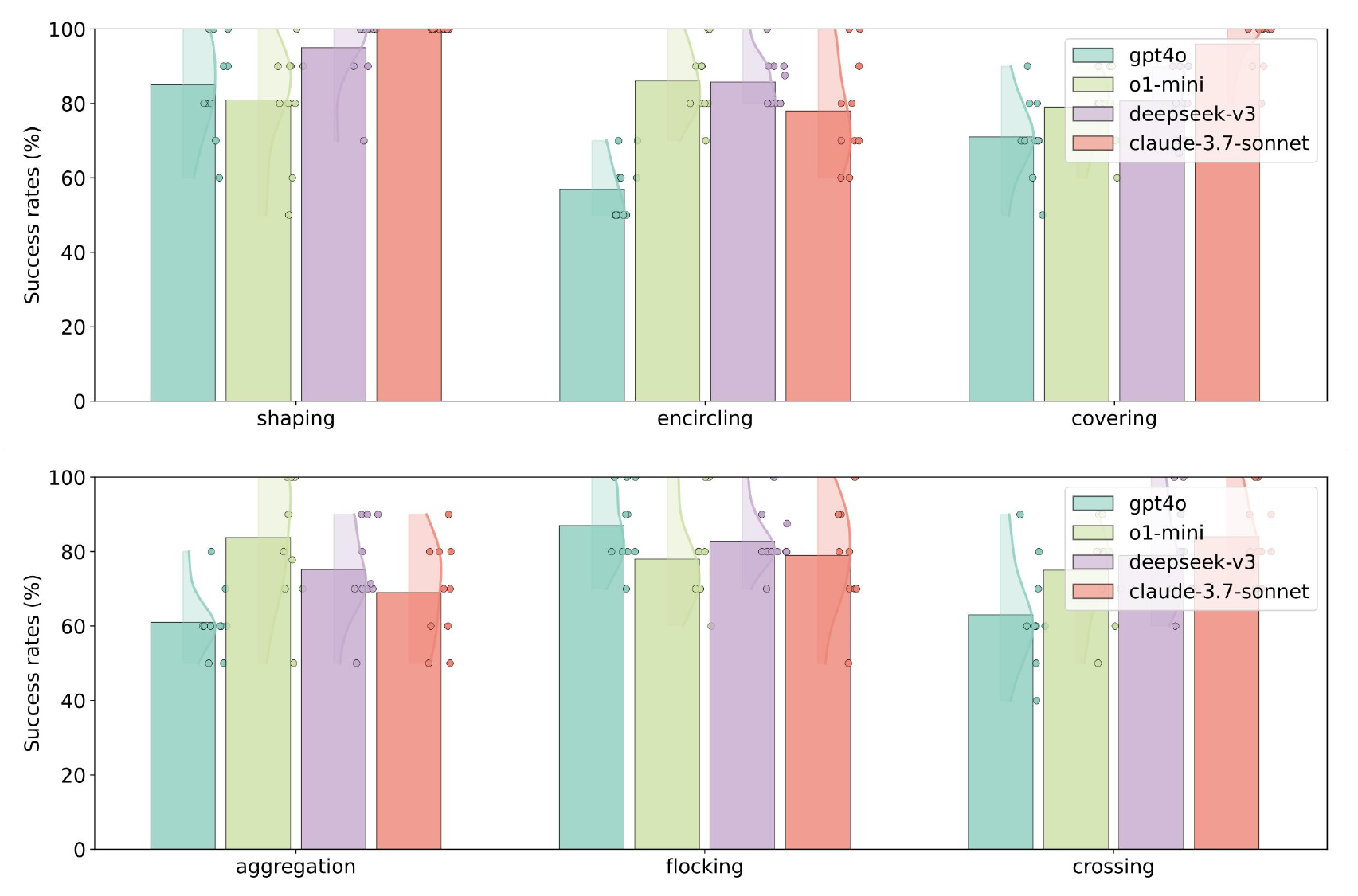}
    \caption{
    \textbf{Success rate comparison between different LLMs.}
    The comparison of four LLM models (GPT-4o, o1-mini, DeepSeek-V3, and Claude-3.7-Sonnet) on multi‑robot task success rates across six representative tasks (shaping, encircling, covering, aggregation, flocking, and crossing). For each combination of model and task, one hundred independent trials were conducted, from user instruction input to policy generation and execution. The average success rates across all tasks for the four models were 71\%, 80\%, 83\%, and 84\%, respectively. Although there are minor performance differences, all models perform reasonably robust across the tasks.}
    \label{different_models}
\end{figure}

\clearpage
\begin{figure}[p]
    \centering
        \includegraphics[width=\linewidth]{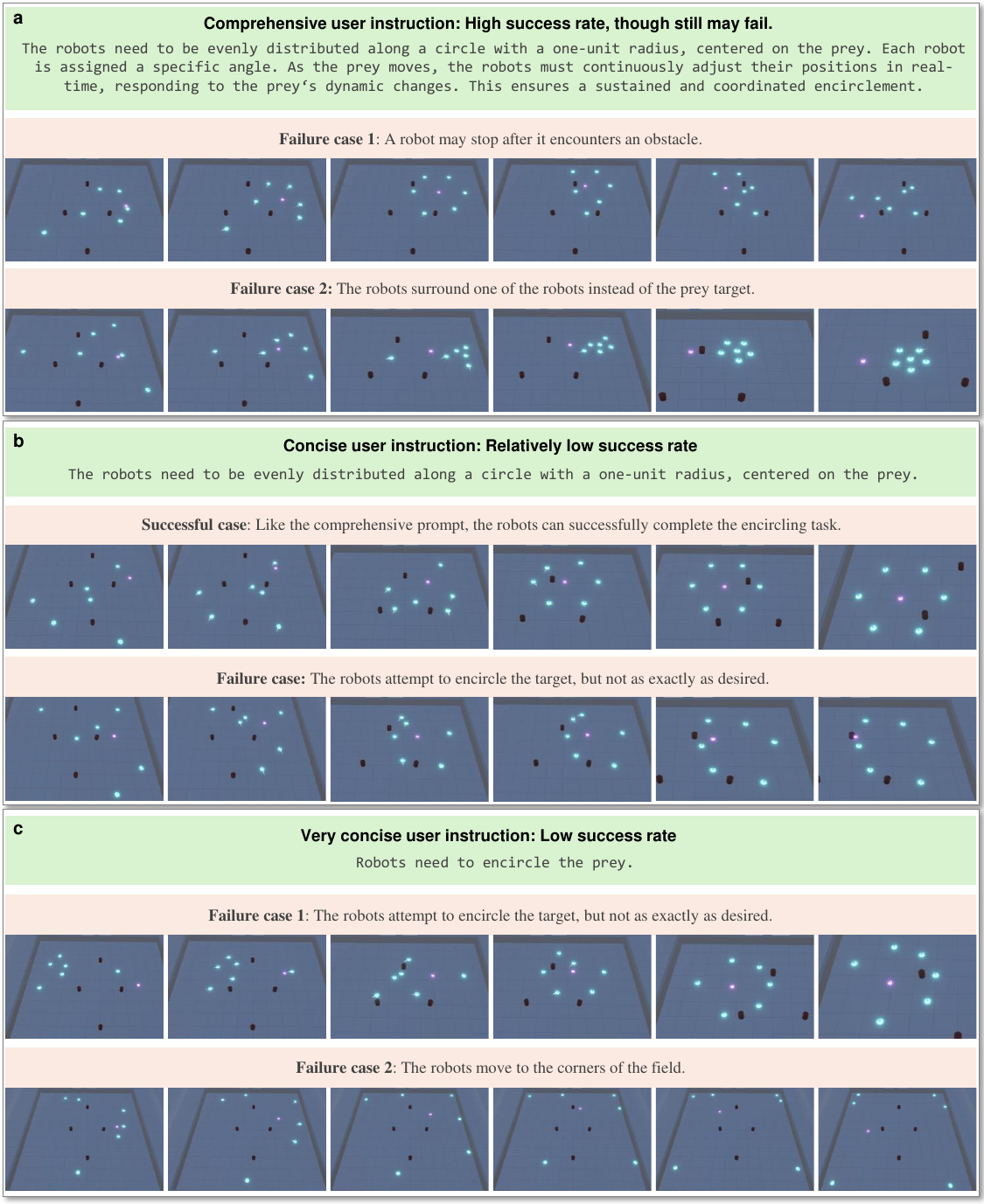}
    \caption{\textbf{Different user instructions for the encircling task and some failure cases.} \textbf{a.} Comprehensive user instructions result in a relatively high success rate yet are not guaranteed to succeed.
    \textbf{b.} Concise user instructions result in a relatively low success rate yet may successfully accomplish the task. \textbf{c.} Overly brief user instructions usually result in a low success rate.}
    \label{fig_failureExample}
\end{figure}

\clearpage
\begin{figure}[p]
    \centering
    \captionsetup{}
    \includegraphics[width=\linewidth]{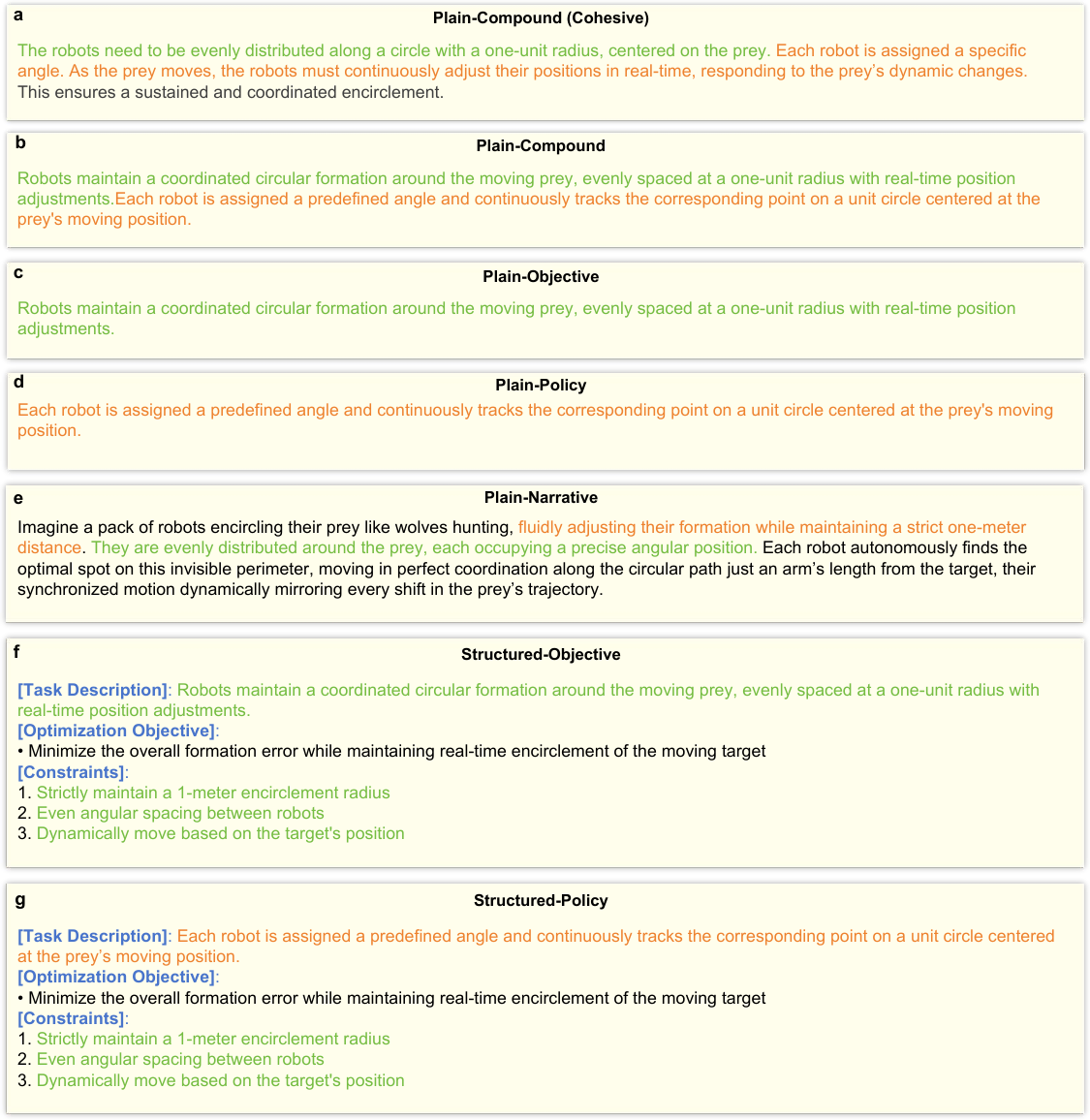}
   \caption{
    \textbf{Examples of seven types of prompts for an encircling task.} We designed seven prompt types, categorized from unstructured to highly structured. In the figure, green text represents task requirements (objectives and constraints), while orange text represents policy instructions.
    \textbf{a.}~Plain-Compound (Cohesive), which integrates both the task objective and policy into a linguistically coherent paragraph.
    \textbf{b.}~Plain-Compound, which concatenates the objective from Plain-Objective and the policy from Plain-Policy without further linguistic integration.
    \textbf{c.}~Plain-Objective, which provides only the task objective without a policy.
    \textbf{d.}~Plain-Policy, which provides only the policy without the task objective.
    \textbf{e.}~Plain-Narrative, which uses natural, narrative language to describe the task.
    \textbf{f.}~Structured-Objective, which restructures the task requirements into a ``description-goal-constraint'' format, with no policy.
    \textbf{g.}~Structured-Policy, which explicitly adds policy information within a structured format.
    }
    \label{encircling_prompt_types}
\end{figure}

\clearpage
\begin{figure}[p]
    \captionsetup{}
    \centering
    \includegraphics[width=1\linewidth]{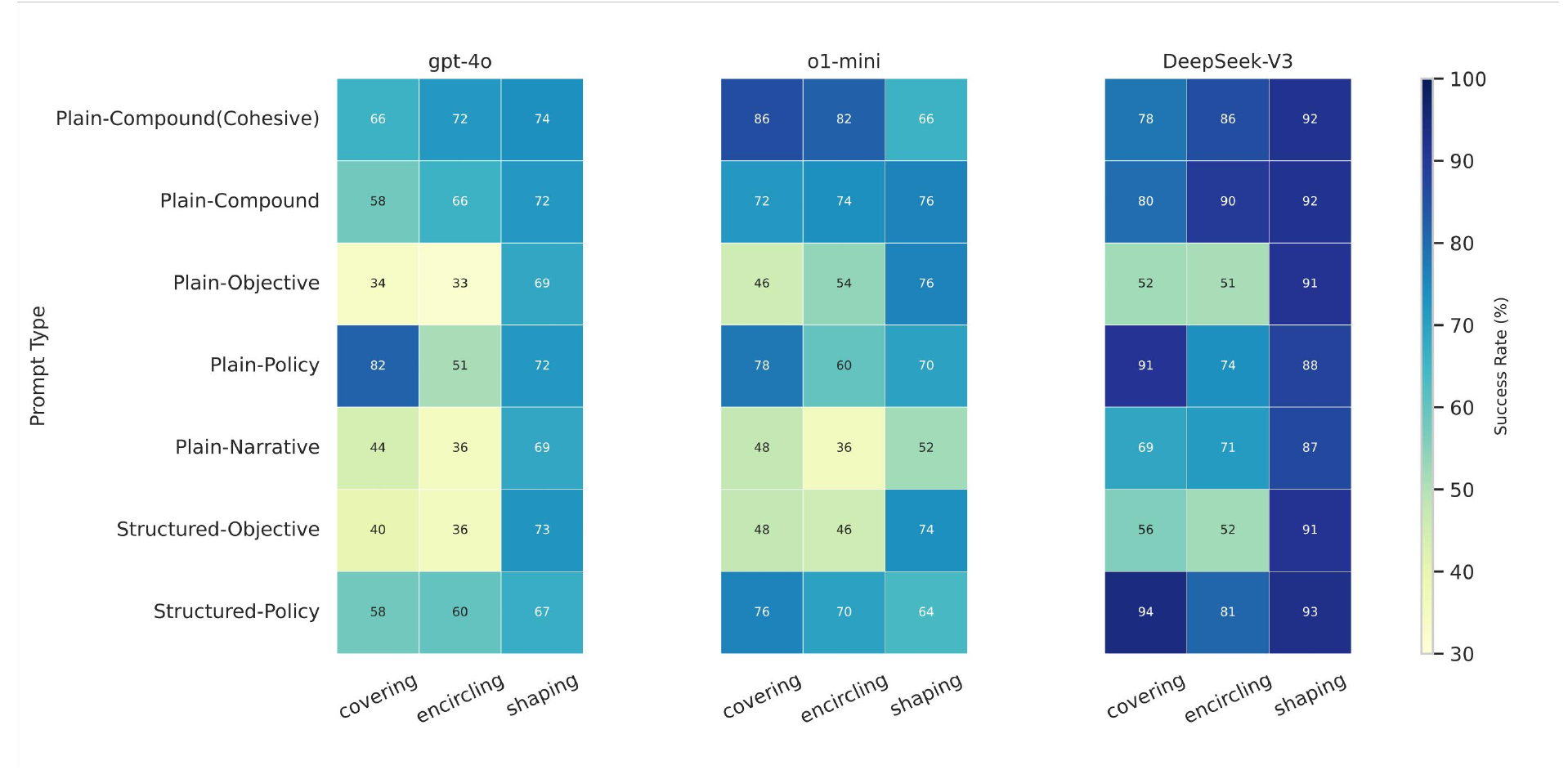}
    \caption{
    \textbf{Comparison between different prompt types.}
    The comparison of seven representative prompt types on multi-robot task success rates across three LLM models (GPT-4o, o1-mini, and DeepSeek-V3) and three representative tasks (covering, encircling, shaping). For each combination of prompt type, model, and task, fifty independent trials were conducted, from user instruction input to policy generation and execution. It is observed that prompt types that contained policy details—{Plain-Compound(Cohesive)} (78\%), {Plain-Compound} (74\%), {Plain-Policy} (74\%), and {Structured-Policy} (74\%)—yielded higher success rates. Conversely, prompts lacking this information, such as {Plain-Objective} (56\%), {Plain-Narrative} (57\%), and {Structured-Objective} (57\%), resulted in lower success rates.
    }
    \label{different_prompts}
\end{figure}

\clearpage
\begin{figure}[p]
    \centering
    \includegraphics[width=\linewidth]{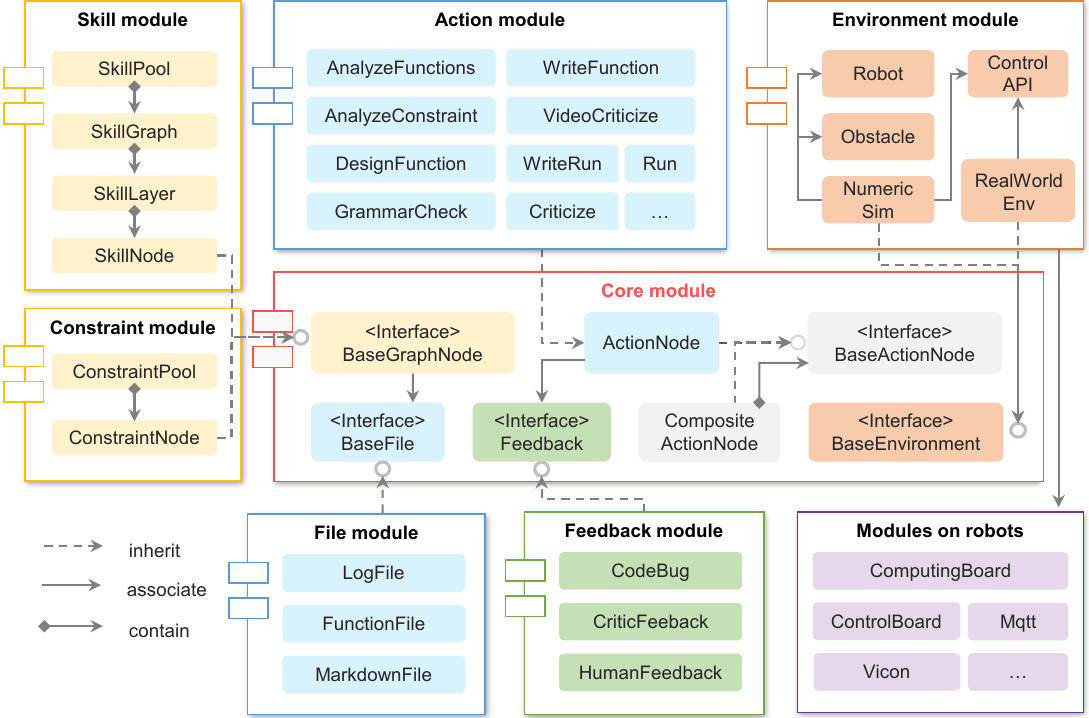}
    \caption{
       \textbf{The software class diagram of GenSwarm consists of seven modules.} The Core Module defines the interfaces between all modules, ensuring seamless integration. The Skill Module manages the skill library and the construction and modification of skill graphs. The Action Module contains action nodes responsible for tasks. The Environment Module supports various simulation and real-world environments, whereas the Constraint Module handles constraint-related tasks. The File Module manages file storage, and the Feedback Module processes all feedback. These modules interact through standardized interfaces defined by the Core Module, ensuring flexibility and consistency across the system.
     }
    \label{fig_softwareClass}
\end{figure}

\clearpage

\begin{figure}[p]
    \centering
    \captionsetup{labelfont={}}
    \includegraphics[width=\linewidth]{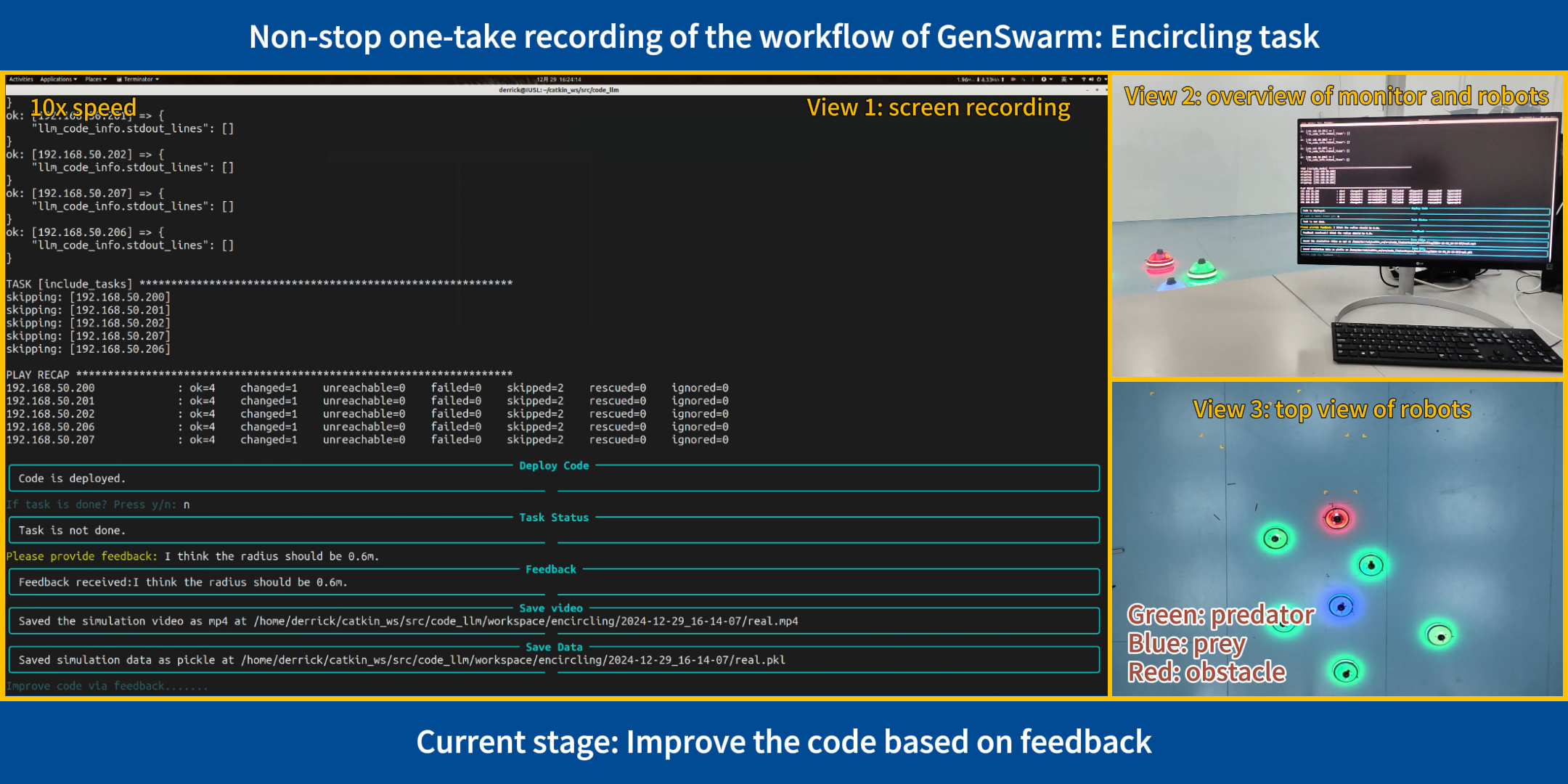}
    \caption{
    \textbf{Screenshot of an experimental video (Movie 1).} \textbf{View 1} shows a screen recording of the computer's terminal, where real-time logs are displayed and a user provides natural language inputs. \textbf{View 2} shows an overview of the physical setup, including the user station and the multi-robot platform. \textbf{View 3} shows a top-down view of the robots executing the task.
    }
    \label{fig_workflow_screenshot}
\end{figure}

\clearpage

\section*{Supplementary Movies}

\href{https://drive.google.com/file/d/11wx835FaOci0608ZwRUVpenIpW6NvKv2/view?usp=drive_link}{\textcolor{blue}{Supplementary Movie 1.}} One-take end-to-end demonstration of GenSwarm performing the encircling task.
\\
\href{https://drive.google.com/file/d/1p3vcjwGjL__qlOe6-OTiutQFwOgGHxow/view?usp=drive_link}{\textcolor{blue}{Supplementary Movie 2.}} One-take end-to-end demonstration of GenSwarm performing the flocking task.

\end{CJK} 
\end{document}